\documentclass{article}

\usepackage{microtype}
\usepackage{graphicx}
\usepackage{subfigure}
\usepackage{booktabs}

\usepackage{hyperref}

\usepackage[accepted]{icml2020}

\icmltitlerunning{Sentiment Analysis in Drug Reviews}

\begin{document}

\twocolumn[
\icmltitle{Sentiment Analysis in Drug Reviews using Supervised Machine Learning Algorithms}

\icmlsetsymbol{equal}{*}

\begin{icmlauthorlist}
\icmlauthor{Sairamvinay Vijayaraghavan}{}
\icmlauthor{Debraj Basu}{}
\end{icmlauthorlist}

\icmlkeywords{Machine Learning, ICML}

\vskip 0.3in
]

\begin{abstract}
Sentiment Analysis is an important algorithm in Natural Language Processing which is used to detect sentiment within some text. In our project, we had chosen to work on analyzing reviews of various drugs which have been reviewed in form of texts and have also been given a rating on a scale from 1-10. We had obtained this data set from UCI machine learning repository which had 2 data sets: train and test (split as 75-25\%). We had split the number rating for the drug into three classes in general: positive (7-10), negative (1-4) or neutral(4-7). There are multiple reviews for the drugs which belong to the similar condition and we decided to investigate how the reviews for different conditions use different words impact the ratings of the drugs. Our intention was mainly to implement supervised machine learning classification algorithms which predicts the class of the rating using the textual review. We had primarily implemented different embeddings such as Term Frequency Inverse Document Frequency (TFIDF) and the Count Vectors (CV). We had trained models on the most popular conditions such as "Birth Control", "Depression" and "Pain" within the data set and obtained good results while predicting on the test data sets.

\end{abstract}

\section{Introduction}
\label{submission}

NLP plays a very vital role in machine learning industry. In particular, it plays a heavy role in using it in the field of medical healthcare through the analysis of medical reviews and texts. There is always an unending demand for a flawless machine learning algorithm which provides absolutely zero error in its predictions. In our study, we had chosen to use a data set which is more on par with the medical healthcare industry and we wanted to investigate the use of NLP algorithms (in particular Sentiment Analysis). 

In general, human beings use sarcasm to imply different meaning by using a completely opposite term in a different connotation. NLP can ensure a model which can be able to understand all these meanings and finally be able to predict the real sentiment intended by the reviewer. An accurate drug review can be classified by the words used contextually within the review and the sentiment present within the review. In our case, we wanted to investigate how crucial are the words used in the review and how do they influence the sentiment prediction of the review and be able to predict the rating of the reviewer. 

Our main goal of this project was to check the effectiveness of using Sentiment Analysis which could detect the sentiment of the review and hence be in agreement with the rating classification. We strongly hypothesized that the words present within the context can play a vital role in determining the sentiment within the review. Natural Language Processing was the key to solving our problem of sentiment detection. Many NLP algorithms and state-of-the-art machine learning models were used for solving this classification problem and the features used were was by converting textual data into numeric data.
\section{Related Work}
There has been a lot of background in using sentiment analysis using machine learning and in particular deep learning algorithms. Emojis have been analyzed and trained for classifying sentiments within tweets (1) and the algorithms used for training these models are: SVM, Naive Bayes, RNN and ANNs. The emojis have been converted into a score using GloVe representation and they have then been used for sentiment detection.

There has also been work done to analyze whether supervised or unsupervised corpus is a better model for predicting sentiments within Tweets and the main deep learning model used are the word embeddings using unsupervised neural language model followed by CNN and then word embeddings using the previous layer are initialized in the next stage which makes it a supervised corpus stage and the results have deduced that supervised model performs better than unsupervised models (2).

There are approaches involving CNN as the main algorithm with a double task prediction of the language of the tweet and the sentiment of the tweet (3).

There has been similar approaches to our work which uses words and their contextual meanings to classify the sentiments within texts and that work focused mainly on the imminent comparison between human based classification and computer based classification (4). There has also been attention scores used for encoding word matrices for sentiments in segment of words (5). However, our work uses mainly numeric representations of the words within the texts (TFIDF and CountVectorizer). 

Ngrams and POS tags have also been used as the main features for classifying tweets and Boosting algorithms were the classifiers for this problem (6,7). The POS tag counts and also the Ngrams were ranked using metrics such as ChiSquared or even in some cases, certain Ngrams were used as a binary feature for marking the presence of certain bigrams.In a particular approach (8), there has been an analysis on high POS tag presence (in particular Adjective) which is used to distinguish product reviews from social reviews while verbs dominate within social reviews than product reviews.

Our approach to sentiment analysis was the contextual meaning of certain key words and how important are certain words in sentiment detection. We had decided to use the word counts (Count Vectorizer) and importance of word in the document (TFIDF scores) as the key features to perform sentiment detection which uses a different form of embedding of words into numeric features and then we decided to compare the performance of neural networks such as ANN and RNN vs the regular classifying state-of-the-art machine learning algorithms such as SVM, Logistic Regression and Random Forests. We decided to investigate a double analysis amongst different algorithms for training models and different feature representation for encoding our texts. To our best knowledge, we have attempted a fresh approach of different combination of algorithms with a slightly different approach of pre-training algorithms for numeric representation of textual data.

\section{Implementation}

For this project, we intended to analyze the words present within a review and how they impact the ratings. First we had treated this problem as a basic NLP problem where we had to classify texts. We had treated this problem as a multi-classification problem which is used to predict the sentiment within the review. Initially, we had to clean up the unclean raw textual data and also we had to reduce the regression problem of predicting numeric ratings into a classification problem of predicting sentiment labels of the reviews. We had then converted our cleaned text data into numeric representations using two key algorithms: Term Frequency Inverse Document Frequency matrix and also the Count Vectorizer matrix.  These algorithms were used with a primary notion of preserving the word count and their contextual meaning within the review. Then, we had used a combination of few supervised machine learning algorithms and also some deep learning algorithms to predict the sentiment label within the review. The algorithms used were Artificial Neural Networks, Recurrent Neural Networks (LSTM and GRU), Support Vector Machines, Random Forests, and Logistic Regression.

\begin{center}
    \includegraphics[width=\columnwidth]{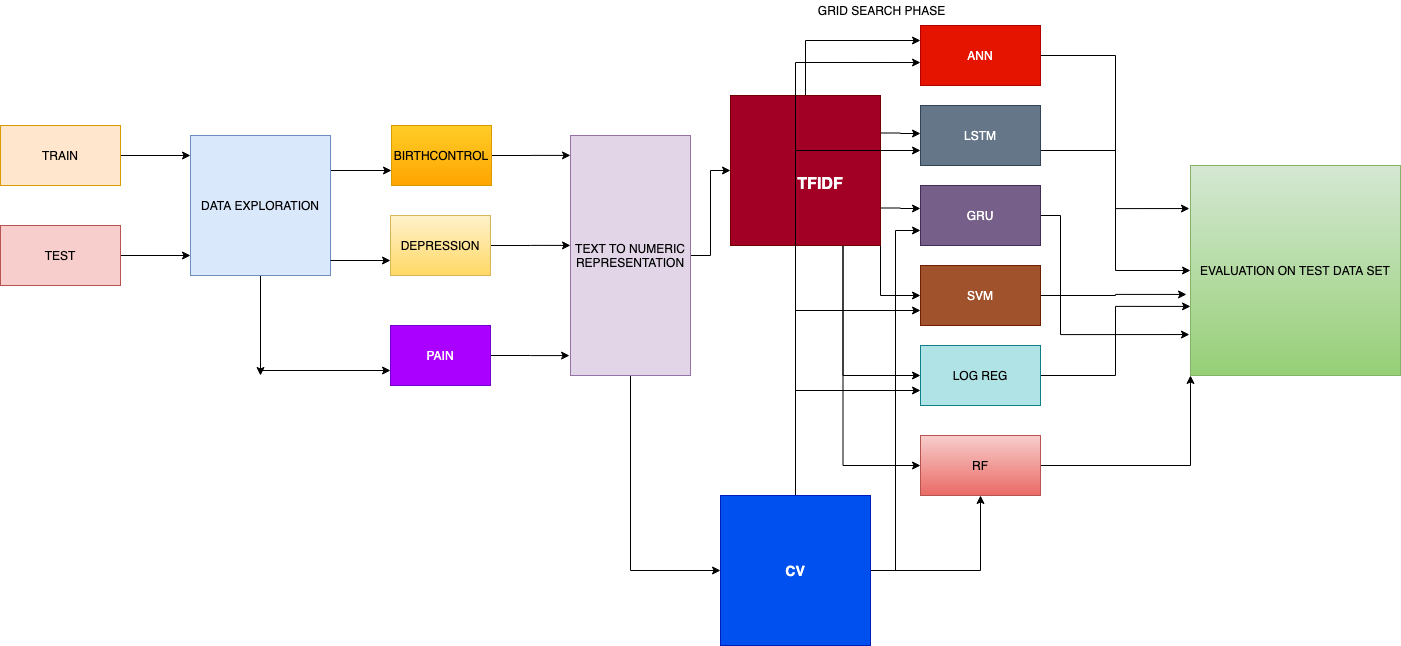}
    \textbf{Pipeline of the Project}
\end{center}

\section{Methodologies}

We had approached this problem in four different phases:

\subsection{PHASE 1: DATA EXPLORATION}

Initially, we analyzed the data set we had used in particular. We had chosen the dataset from the UCI machine learning repository. The entire data has been rendered in two files: train and test tsv files. The train and test files had a train test split of (75-25\%) the number of samples. The entire data set had around 215063 number of samples. Each sample has the following fields: the Drug Name, condition for which drug is used, review by an user for this drug, the rating given by the user, the date on which drug was reviewed, and the useful\_Count which represents number of users who found the count useful. 

For this project, we entirely focused on testing the behavior of the textual data and how the presence of words within each review plays a role in predicting the sentiment of the review. We had first worked on exploring more about the text within the review. We decided to focus on 3 different conditions and wanted to analyze how to review for each condition varies in sentiment detection. We had split the number ratings for the particular drug from a range of (1-10) into three classes in general: positive (7-10), negative (1-4) or neutral(4-7) to convert the regression problem into a classification problem.

In order to analyze the reviews, we trained and tested only on the reviews of 3 most popular condition: Birth Control, Depression and Pain. 

We chose to do within these conditions since the entire data set had an uneven distribution of ratings with 60\% positive ratings, 25\% negative ratings and 15\% neutral ratings. So, we wanted to narrow the distribution to smaller number of samples and hence we chose to work within the different conditions specifically.
\begin{figure}[h]
    \centering
\includegraphics[scale = 0.45]{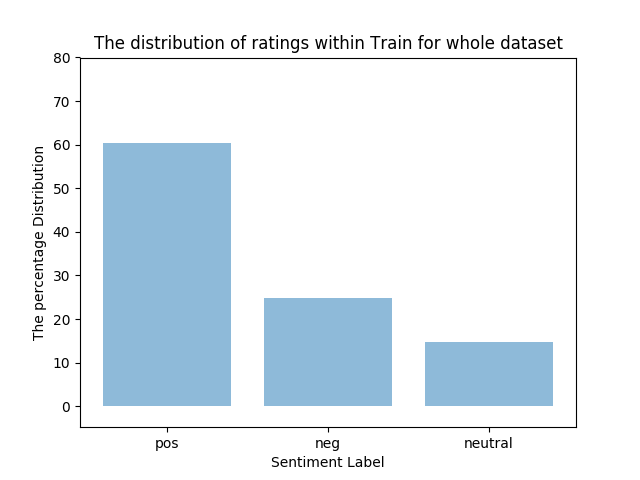}\\
\includegraphics[scale = 0.45]{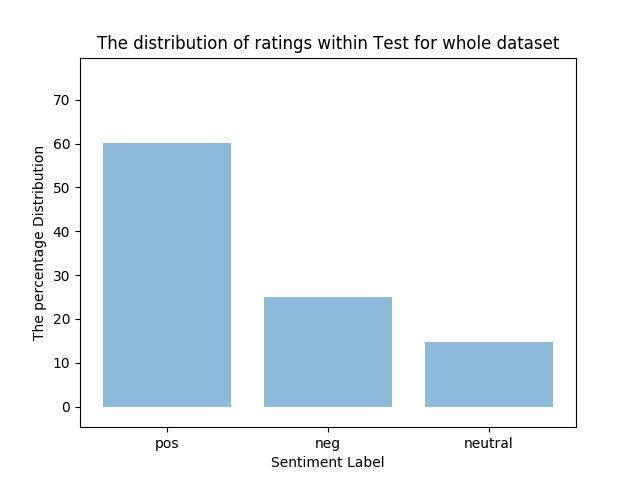}
\end{figure} 

We had also analyzed the tag distributions for the entire data set and we found that the tag distribution is not very helpful to classify the sentiment within review since we find a similar pattern of high number of Adjectives and Adverbs within the positive reviews in comparison of all the other reviews. However, we decided not to use this feature since we found that there is a given imbalance of many reviews and the adjective and adverb proportion follow a similar proportion to that of the distributions which means that since there are more positive reviews, there are more words for positive reviews and hence there is a predominance of adverbs and adjectives. Hence we chose to investigate the words within the review in order to classify the sentiments.

\includegraphics[scale= 0.45]{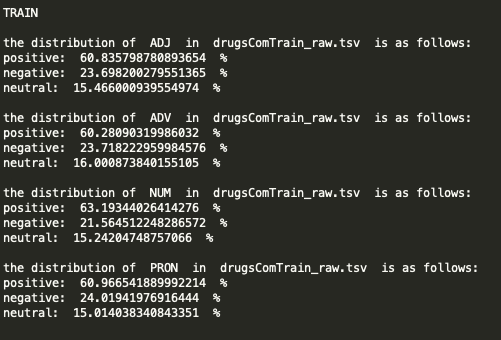}\\
\includegraphics[scale = 0.45]{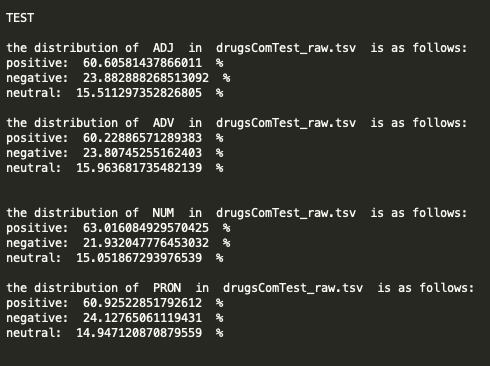}

We found that there was a lot of common words present in every review which have no significant meaning such as determinants,pronouns within the reviews, we chose to do a simple data cleanup of the texts by converting the texts to lowercase and also remove the digits and the commonly present unimportant words such as "a","an","the","them", etc. and also the punctuation characters present in the data set. We also removed the non-alpha characters within the data set so that we get a cleaned data set with non-redundant and more significant features for analysis.

Hence, after we obtained the cleaned up text data, we decided to investigate the importance of certain words to influence the sentiment within the words.\\\\

\begin{table}[]
    \centering
    \begin{tabular}{c|c|c|c}
    
    Condition  &  Train & Test & Num features (Different Words)\\
    \hline
    Birth Control & 28788 & 9648 & 13311\\
    \hline
     Depression    & 9069& 3095 & 9692\\
     \hline
     Pain & 6145 & 2100 & 7686
    \end{tabular}
    \caption{Number of Samples and Features for TFIDF and CV for Each Condition}
\end{table}

\subsection{PHASE 2: Text to Numeric data Representation}
In order to encode the review texts into numeric data, we had used certain pre-training algorithms such as Term Frequency Inverse Document Frequency and also Count Vectorizer. The reason behind using these algorithms is because we wanted to encode the importance of the presence of certain words and these algorithms encode the importance and presence of words in different ways. We used Term frequency-Inverse Document Frequency (TF-IDF) embedding to calculate the matrix of numeric values for each word $t$ within each review texts. If,
The term frequency $tf(t, d)$ calculates the proportion of times that the term $t\in V(d)$ appears in the document $d$. The vocabulary $V(d) = \sum_t n(t,d)$ is constructed by the document $d$. Thus, if a word $w'$ does not appear in a document $d'$, the term frequency $tf(t', d')$ in this case would be zero. The idea of the term frequency is essentially the same as CountVectorizer.

$$tf(t,d) = \frac{n(t,d)}{V(d)}$$
$$n(t,d) = \textrm{ occurrence of the word }t\textrm{ in the document }d$$

Given a document collection $D$, the inverse document frequency $idf(t, D)$ is the log of the number of documents $N$ divided by $df(t,D)$, the number of documents $d \in D$ containing the term $t$. As a result, common words in $D$ will have a low term frequency score, while infrequent words will have a high term frequency. Thus, the term frequency will be very likely to separate fake news that often have less common words (even ungrammatical) from real news that usually consist of common words.
$$idf(t,D) =\log \Big(\frac{N}{df(t,D)}\Big)$$

As a summary, TF-IDF score $w(t,d)$ for a word increases with its count, but will be counteracted if the word appears in too many documents. 
$$w(t,d) = tf(t,d) \times idf(t, D)$$

Similarly, count vectorizer is given by a matrix of values with each value representing the count frequency of that word within that document (review). This matrix is the one hot encoded representation of the different words present in the corpus. Entry $a_{ij} =$ total number of times $j$th word appears in the $i$th document.

We did this for each of the different conditions separately since we wanted to verify how would words within different condition play a major role in sentiment analysis.

\subsection{PHASE 3: Training Models}
We had decided to try these following algorithms to investigate the accuracy of sentiment detection using the above numeric representation techniques. We used algorithms such as Neural Networks: Artificial Neural Networks (ANN), Recurrent Neural Networks with Long Short Term Memory (LSTM) and Gated Recurrent Units (GRU) and also other state-of-the-art machine learning classification algorithms such as: Support Vector Machines (SVM), Logistic Regression (LR) and also Random Forests (RF). Our next aim was to identify which type of machine learning algorithm would yield best results. We had trained 6 * 2 (for each encoding) = 12 models for each condition (3 conditions we had trained our model for).

We had performed grid search for each algorithm which was performed to obtain the best hyper parameters for each model in order to get the best results.

We had done a 10 Fold Evaluation for each of the Grid Search models and we picked the best model based on the best accuracy. After picking the best hyper parameters, we then evaluated our model on the test data set reviews for that corresponding condition alone (that is if trained on the reviews from Birth Control condition from the train data set, we tested on the reviews from Birth Control condition from the test data.

\subsection{PHASE 4: NOT INCLUDED BUT WHAT WE EXPERIMENTED: tried the W2V and also the different condition predict using BC}

In addition to the other numeric representations, we had investigated the use of W2V embeddings on our review texts by using a word2vec model which is also another version of encoding textual data into numeric data by containing contextual information about words. We had trained our Word2Vec on the review texts itself and applied the algorithms on the data. However, we decided not to proceed with this approach since we achieved a very poor accuracy of 67\% of testing and training accuracy and hence we did not proceed with this model. We attribute this failure to the cause that Word2Vec is not that effective in capturing word importance within a context and this is because the texts we had provided to the Word2Vec model is not enough to learn about its contextual meaning.

\section{Results}

NOTE: We denoted Random Forest as RF, Artificial Neural Network as ANN, Logistic Regression as LogRegr, Long Short Term Memory as LSTM, Gated Recurrent Unit as GRU, Support Vector Machines as SVM

BCE stands for Binary Cross-Entropy Loss function
\subsection{Grid Result Table}
The following tables represent the summary of our grid search results for each of the models for the 3 most popular condition Birth Control, Depression and Pain:\\\\\\\\\\\\\\\\\\

\vline
\normalsize
\begin{tabular}{lll|}
\toprule
 & TFIDF & CountVectorizer \\
\hline
 ANN &activation = relu&activation = relu\\
        &optimizer = RMSProp & optimizer = Nadam\\
        &hidden\_layers = 2 & hidden\_layers = 2\\
        &hidden\_neurons = 400 & hidden\_neurons = 400\\ 
        &loss = BCE & loss = BCE\\
        \hline
        LSTM &activation = linear & activation = linear\\
        &optimizer =Adam & optimizer = Nadam\\
        &hidden\_layers = 2 & hidden\_layers = 2\\
        &hidden\_neurons = 500&hidden\_neurons = 500\\
        &mem\_cells = 300&mem\_cells=400 \\
        &loss = BCE & loss = BCE\\
        
        \hline
        GRU&activation = softsign & activation = linear\\
        &optimizer =Adam & optimizer = Nadam\\
        &hidden\_layers = 4 & hidden\_layers = 3\\
        &hidden\_neurons = 200&hidden\_neurons = 500\\
        &mem\_cells = 400 &mem\_cells=200 \\
        &loss = BCE & loss = BCE\\
        \hline
        SVM&C=1, Kernel = ‘linear’&C=1, Kernel = ‘linear’\\
        \hline
        LogRegr&C=1, penalty=L2&C=1, penalty=L2\\
        \hline
        RF&max\_depth=9&max\_depth=9\\
&max\_features=sqrt&max\_features=sqrt\\
&min\_samples\_leaf=4&min\_samples\_leaf=3\\
&min\_samples\_split=2&min\_samples\_split=8\\
&num\_trees=200&num\_trees=200\\
         \hline
\end{tabular}
\textit{Hyperparameter search: Condition: Birth Control}

\vline
\normalsize
\begin{tabular}{lll|}
\toprule
 & TFIDF & CountVectorizer \\
\hline
ANN &activation = softsign&activation = relu\\
        &optimizer = Adam & optimizer = Nadam\\
        &hidden\_layers = 2 & hidden\_layers = 3\\
        &hidden\_neurons = 400 & hidden\_neurons = 400\\ 
        &loss = BCE & loss = BCE\\
        \hline
        LSTM &activation = softsign & activation = linear\\
        &optimizer =Nadam & optimizer = Nadam\\
        &hidden\_layers = 2 & hidden\_layers = 3\\
        &hidden\_neurons = 200&hidden\_neurons = 300\\
        &mem\_cells = 200&mem\_cells=500 \\
        &loss = BCE & loss = BCE\\
        
        \hline
        GRU&activation = softsign & activation = relu\\
        &optimizer =Nadam & optimizer = Nadam\\
        &hidden\_layers = 2 & hidden\_layers = 2\\
        &hidden\_neurons = 500&hidden\_neurons = 200\\
        &mem\_cells = 300&mem\_cells=300 \\
        &loss = BCE & loss = BCE\\
        \hline
        SVM&C=1, Kernel = ‘linear’&C=1, Kernel = ‘linear’\\
        \hline
        LogRegr&C=1, penalty=L2&C=1, penalty=L2\\
        \hline
        RF&max\_depth=1&max\_depth=1\\
&max\_features=auto&max\_features=auto\\
&min\_samples\_leaf=3&min\_samples\_leaf=3\\
&min\_samples\_split=2&min\_samples\_split=2\\
&num\_trees=200&num\_trees=200\\
         \hline
\end{tabular}
\textit{Hyperparameter search: Condition: Depression}

\normalsize
\vline
\begin{tabular}{lll|}
\toprule
 & TFIDF & CountVectorizer \\
\hline
ANN &activation = linear&activation = relu\\
        &optimizer = Adam & optimizer = Nadam\\
        &hidden\_layers = 2 & hidden\_layers = 3\\
        &hidden\_neurons = 200 & hidden\_neurons = 500\\ 
        &loss = BCE & loss = BCE\\
        \hline
        LSTM &activation = tanh & activation = tanh\\
        &optimizer =Nadam & optimizer = Nadam\\
        &hidden\_layers = 4 & hidden\_layers = 2\\
        &hidden\_neurons = 200&hidden\_neurons = 200\\
        &mem\_cells = 300&mem\_cells=200 \\
        &loss = BCE & loss = BCE\\
        
        \hline
        GRU&activation = tanh & activation = linear\\
        &optimizer =Nadam & optimizer = Nadam\\
        &hidden\_layers = 4 & hidden\_layers = 3\\
        &hidden\_neurons = 200&hidden\_neurons = 400\\
        &mem\_cells = 500&mem\_cells=400 \\
        &loss = BCE & loss = BCE\\
        
        \hline
        SVM&C=1, Kernel = ‘linear’&C=1, Kernel = ‘linear’\\
        \hline
        LogRegr&C=1, penalty=L2&C=1, penalty=L2\\
        \hline
        RF&max\_depth=1&max\_depth=1\\
&max\_features=auto&max\_features=auto\\
&min\_samples\_leaf=3&min\_samples\_leaf=3\\
&min\_samples\_split=2&min\_samples\_split=2\\
&num\_trees=200&num\_trees=200\\
         \hline
\end{tabular}
\large \textit{Hyperparameter search: Condition: Pain}

\newpage
\subsection{Accuracy Result Tables}
\begin{table}[]
    \centering
    \begin{tabular}{|c|c|c|}
         Algorithm& train K=10 fold &test data \%accuracy\\
         \hline
         ANN&92.5142 & 93.4114\\
         \hline
         LSTM&87.9915 & 88.9373\\
          \hline
         GRU&88.1397 & 88.3707\\
          \hline
         SVM&82.2183& 75.2177\\
          \hline
         LogReg&77.4281 & 73.2794\\
          \hline
         RF&54.8666&53.4515\\
          \hline
         
    \end{tabular}
    \caption{Accuracy for each best model: TFIDF feature and Condition: Birth Control}
    \label{tab:my_label}
\end{table}

\begin{table}[]
    \centering
    \begin{tabular}{|c|c|c|}
         Algorithm& train K=10 fold &test data \%accuracy\\
         \hline
         ANN&92.7898 & 93.85\\
         \hline
         LSTM&91.9226 & 93.2145\\
          \hline
         GRU&90.8747 & 92.4475\\
          \hline
         SVM&91.6111 & 80.369\\
          \hline
         LogReg&86.8348 & 77.4979\\
          \hline
         RF&54.0503&52.9747\\
          \hline
         
    \end{tabular}
    \caption{Accuracy for each best model: Count Vectorizer feature and Condition: Birth Control}
    \label{tab:my_label}
\end{table}

\begin{table}[]
\centering
\begin{tabular}{|c|c|c|}
Algorithm& train K=10 fold &test data \%accuracy\\
\hline
    ANN &88.7309 &90.1993 \\
\hline
    LSTM & 88.6279& 90.6085\\
\hline
    GRU & 88.4625& 90.0162\\
\hline
   SVM  & 87.4407& 77.6737\\
\hline
   LogReg & 80.1632& 74.2811\\
\hline
   RF  & 61.0872& 61.1632\\
\hline
\end{tabular}
 \caption{Accuracy for each best model: TFIDF feature and Condition: Depression}
    \label{tab:my_label}
\end{table}

\begin{table}[]
\centering
\begin{tabular}{|c|c|c|}
Algorithm& train K=10 fold &test data \%accuracy\\
 \hline
    ANN & 81.3357 & 92.1056 \\
\hline
    LSTM &88.6132 & 90.1669 \\
\hline
    GRU & 88.7272 & 90.0808\\
\hline
   SVM  & 97.3426 & 84.4265\\
\hline
   LogReg & 93.6818 & 82.0032\\
\hline
   RF  & 61.0872 & 61.1632\\
\hline
\end{tabular}
 \caption{Accuracy for each best model: CountVectorizer feature and Condition: Depression}
    \label{tab:my_label}
\end{table}

\begin{table}[]
\centering
\begin{tabular}{|c|c|c|}
Algorithm& train K=10 fold &test data \%accuracy\\
\hline
    ANN & 88.2343 & 89.6667 \\
\hline
    LSTM & 88.3482 & 89.4127\\
\hline
    GRU & 87.9902 & 89.55556\\
\hline
   SVM  & 86.7535 & 79.0476\\
\hline
   LogReg & 79.8373 & 76.5238\\
\hline
   RF  & 70.5940 & 70.2381\\
\hline
\end{tabular}
 \caption{Accuracy for each best model: TFIDF feature and Condition: Pain}
    \label{tab:my_label}
\end{table}

\begin{table}[]
\centering
\begin{tabular}{|c|c|c|}
Algorithm& train K=10 fold &test data \%accuracy\\
 \hline
    ANN &  89.99 & 91.286\\
\hline
    LSTM & 88.1964 & 88.7619\\
\hline
    GRU & 87.8004 & 89.0635\\
\hline
   SVM  & 96.9731 & 83.1905\\
\hline
   LogReg & 94.0277 & 83.0476\\
\hline
   RF  & 70.5940 & 70.2381\\
\hline
\end{tabular}
 \caption{Accuracy for each best model: CountVectorizer feature and Condition: Pain}
    \label{tab:my_label}
\end{table}
\newpage

\subsection{Graphs}

Here's the ROC and PR plots with F1 score for each model with their respective AUC scores with each plot representing the algorithm for that model

   
\includegraphics[scale = 0.5]{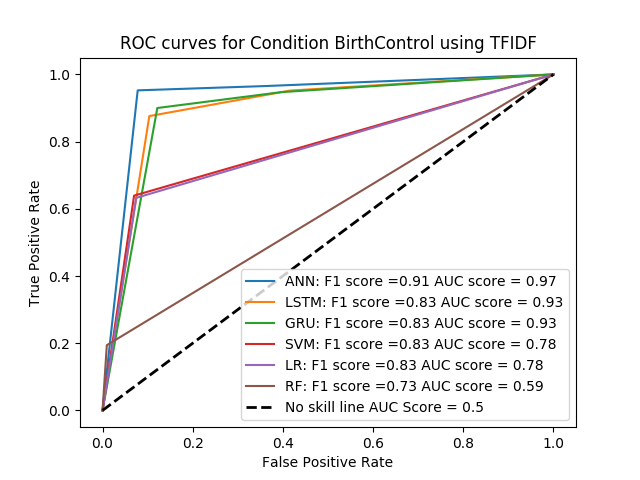}   
\includegraphics[scale = 0.5]{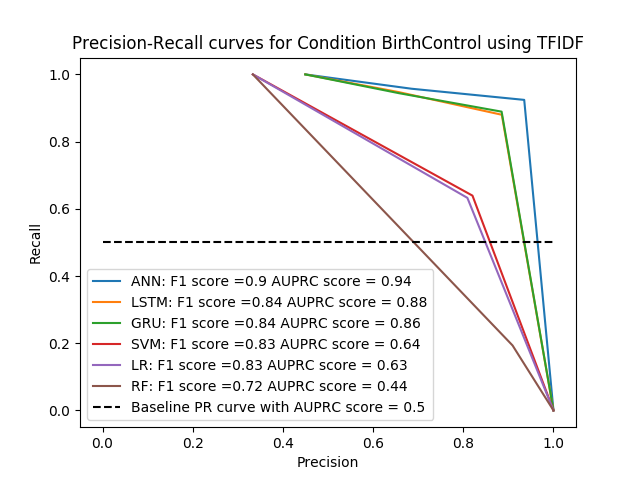}  
\includegraphics[scale = 0.5]{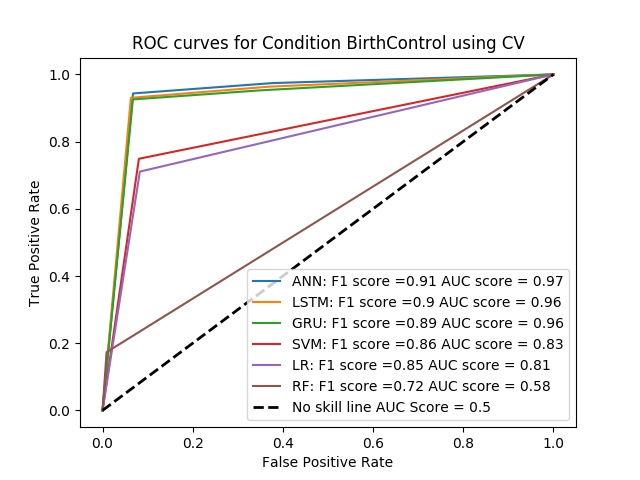}
\includegraphics[scale = 0.5]{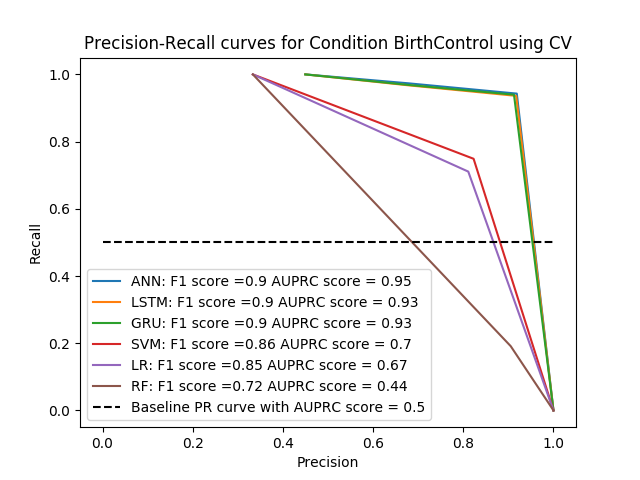}
\includegraphics[scale = 0.5]{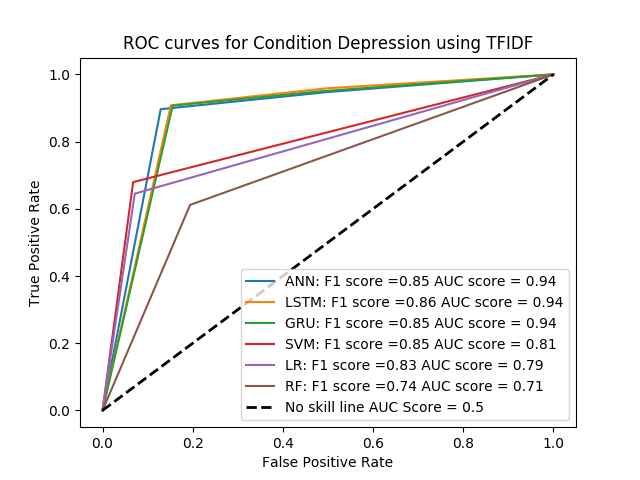}
\includegraphics[scale = 0.5]{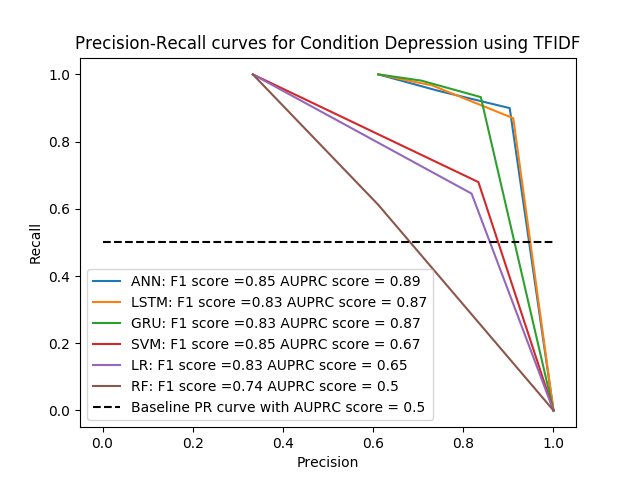}
\includegraphics[scale = 0.5]{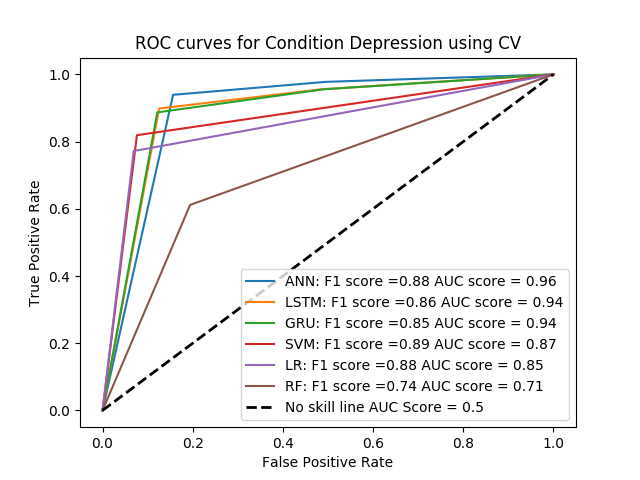}
\includegraphics[scale = 0.5]{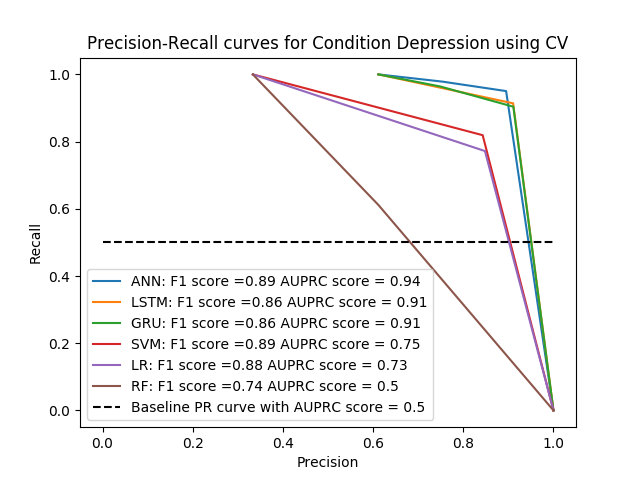}
\includegraphics[scale = 0.5]{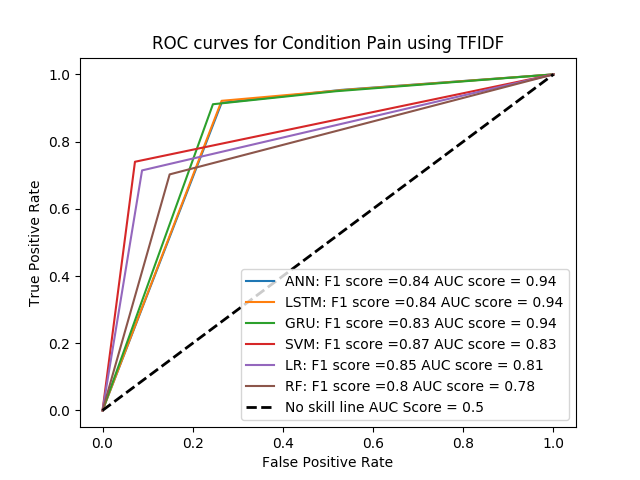}
\includegraphics[scale = 0.5]{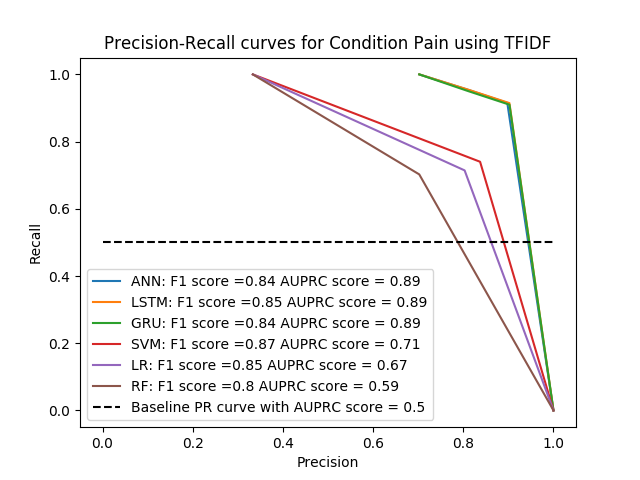}
\includegraphics[scale = 0.5]{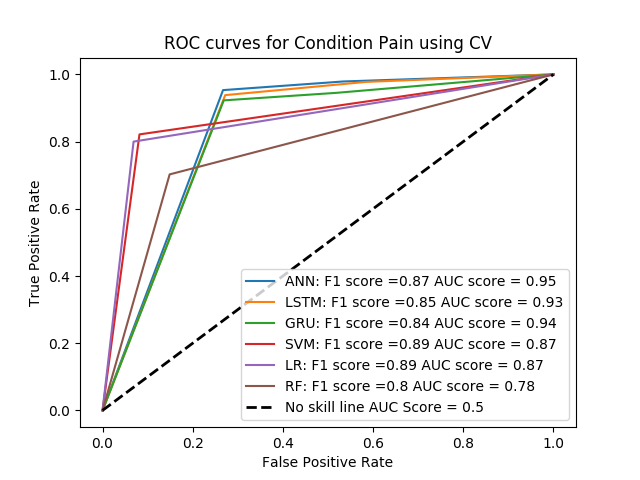}
\includegraphics[scale = 0.5]{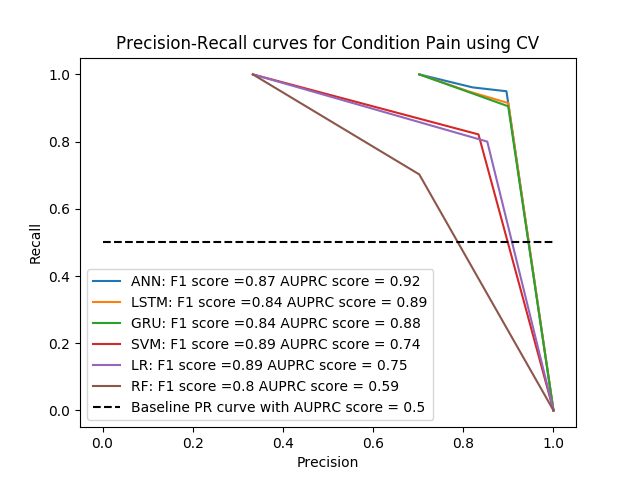}

\newpage

\section{Conclusion}

From our results, we are deducing the following conclusions. We found out a general pattern where Count Vector does better than the TFIDF encodings. The reason why CV does better is because of the blatant openness in representing the words within the reviews. CV represents mainly the count of words present in the reviews while TFIDF represents the significance of the words within the review. Since TFIDF hides the entire contextual meaning of the word within the review, it does worse than the CV encoding. 

Similarly, we found another conclusion from the algorithms. In general, we noted that neural networks in particular did better prediction on test data sets than the other machine learning algorithms. We found this particular for different conditions. The reason why this happens is because we found that deep learning algorithms classify better than the other machine learning algorithms. Deep learning algorithms capture more significant features for classifying to predict the sentiment within the review. This therefore accounts for including all the information crucial for classifying the sentiment within the reviews. Other algorithms don't predict the sentiment within the review as accurate in comparison to the deep learning algorithms because of the same reason. Neural Network use all the features from the  Hence, we conclude that deep learning algorithms with Count Vectorizer encoding do a better job than the other models in general for each condition. With respect to both the RNN models, both the models perform similarly to each other. LSTM and GRU are similar to each other for every condition and hence we can conclude Recurrent Neural Networks have a similar performance amongst each other. 

SVM and Logistic Regression also have a pattern in common. They perform similar in every model with SVMs having an edge over Logistic Regression. The reason why we find it is because SVM can provide a better algorithm for classifying since it does margin classification while Logistic Regression classifies based on probability of likelihood which does not provide to be a better classifier based on the significant features (words presence) used in TFIDF and CV. 

In addition to the above patterns, we noticed that random forest algorithm models in general performed the worst in comparison to the other algorithms. The reason we can deduce for this performance is because random forest is a decision tree classifier which probably fails to include all the crucial features used for classifying unlike neural networks or SVMs.

\section{References}
1.Dane Hankamer, David Liedtka. Twitter Sentiment Analysis with Emojis. 2019\\\\
2.Aliaksei Severyn, Alessandro Moschitti.Twitter Sentiment Analysis with Deep Convolutional Neural Networks. 2015\\\\
3.Jônatas Wehrmann et al. A Multi-Task Neural Network for Multilingual Sentiment Classification and Language Detection on Twitter. 2018\\\\
4.Simon Provoost et al. Validating Automated Sentiment Analysis of Online Cognitive Behavioral Therapy Patient Texts- An Exploratory Study. 2019\\\\
5.Stefanos Angelidis and Mirella Lapata. Multiple Instance Learning Networks for Fine-Grained Sentiment Analysis. 2018\\\\
6.Efthymios Kouloumpis, Theresa Wilson, Johanna Moore. Twitter Sentiment Analysis: The Good the Bad and the OMG!. 2011\\\\
7.Alexander Pak, Patrick Paroubek. Twitter as a Corpus for Sentiment Analysis and Opinion Mining. 2010\\\\
8.Mostafa Karamibek, Ali A. Ghorbani. Sentiment Analysis of Social Issues. 2012\\\\
\end{document}